\documentclass[letterpaper, 10 pt, conference]{ieeeconf}

\IEEEoverridecommandlockouts
\usepackage{booktabs}
\usepackage{tabularx}
\usepackage{wrapfig}
\usepackage{cite}
\usepackage{amsmath,amssymb,amsfonts}
\usepackage{algorithmic}
\usepackage{graphicx}
\usepackage{textcomp}
\usepackage{xcolor}
\usepackage{subcaption}
\usepackage[font=small]{caption}
\usepackage{hyperref}
\usepackage{balance}
\def\BibTeX{{\rm B\kern-.05em{\sc i\kern-.025em b}\kern-.08em
    T\kern-.1667em\lower.7ex\hbox{E}\kern-.125emX}}
\begin{document}

\title{\LARGE \bf Perceptive Locomotion with Controllable Pace\\and Natural Gait Transitions Over Uneven Terrains}

\author{Daniel Chee Hian Tan$^{1*}$, Jenny Zhang$^{1*}$, Michael (Meng Yee) Chuah$^{1}$ and Zhibin Li$^{2}$
\thanks{$^{1}$Authors are with Institute for Infocomm Research, A*STAR, Singapore}%
\thanks{$^{2}$Author is with the Department of Computer Science, University College London, United Kingdom}%
\thanks{$^{*}$Equal contribution}%
}

\bstctlcite{IEEEexample:BSTcontrol}

\maketitle
\thispagestyle{empty}
\pagestyle{empty}
\begin{abstract}
This work developed a learning framework for perceptive legged locomotion that combines visual feedback, proprioceptive information, and active gait regulation of foot-ground contacts. The perception requires only one forward-facing camera to obtain the heightmap, and the active regulation of gait paces and traveling velocity are realized through our formulation of CPG-based high-level imitation of foot-ground contacts. Through this framework, an end-user has the ability to command task-level inputs to control different walking speeds and gait frequencies according to the traversal of different terrains, which enables more reliable negotiation with encountered obstacles. The results demonstrated that the learned perceptive locomotion policy followed task-level control inputs with intended behaviors, and was robust in presence of unseen terrains and external force perturbations. A video demonstration can be found \href{https://youtu.be/OTzlWzDfAe8}{here}, and the codebase \href{https://github.com/jennyzzt/perceptual-locomotion}{here}.
\end{abstract}

\section{Introduction}
Legged locomotion has achieved commendable performance without visual perception of the external environment, which can negotiate certain vertical heights via reactive reflexes \cite{peng_learning_2020, kumar_rapid_2021, peng_amp_2021}. However, perceptive locomotion enables proactive terrain negotiation, anticipating and avoiding obstacles \cite{miki_learning_2022}, augmenting traversability of various terrains. The recent DARPA subterranean challenge \cite{roucek_darpa_2020} exemplified the need for legged robots capable of terrain-adaptive locomotion, in order to achieve better safety and efficiency for mission completion in many real-life applications. Overall, legged robots are becoming commodities and more advantageous compared to their wheeled and tracked counterparts.

The required technologies for doing so boils down to robot perception and control. In well established robotics streams, perception involves engineering the extraction of visual features that can be modularly connected with designed controllers, i.e. integrating SLAM and robot state estimation with the existing blind locomotion modules, such as optimization-based motion planning \cite{eth-non-learning}, multi-skill locomotion, \cite{yang2020multi}, reinforcement learning based trajectory adaptation \cite{Gangapurwala2021}, and meta-reinforcement learning \cite{anne2021meta}.
Despite this modularity and interpretability, traditional control architectures without learning require significant engineering effort and expertise over multiple disciplines for such systems engineering, resulting in longer development cycles. 

In contrast, learning-based approaches can bridge sensing and closed-loop control by directly feeding perceptive data into the control policy. The ability to directly design an end-to-end controller offers better scalability in increasing the dimensionality for accessing both proprioceptive and exteroceptive data within the feedback loop -- a more integral and compact scheme to design a perception-action control system with environment-aware behaviors. 

The real-world challenges in DARPA subterranean finals in September 2021 \cite{roucek_darpa_2020} found that apart from perceiving irregular surfaces, it is equally important to control different gait frequencies at different target traveling velocities, in order to increase the success rate of terrain negotiation. This is specially observed when many robots fell due to keeping an almost constant trotting frequency over different traveling speed on uneven terrains, resulting in them toppling over and mission failures. 

This work aims to develop computationally efficient and effective feedback control for terrain-aware motor skills, excluding the long-term motion planning. Our contributions are: (i) active obstacle negotiation with only one front camera using depth images; (ii) a learning framework with decoupled modules for perception and motor control policy; (iii) controllable gait frequency and velocity as high-level inputs for operators to regulate robot gaits in accordance with different terrains. 

The proposed scheme offers cost-effective solutions for perceptive locomotion, and the straightforwardness of each sub-module makes them easily implementable, because the perception requires only one off-the-shelf depth camera and can be re-trained separately to embed new targeted terrain features, and the computation of a forward neural network (NN) is very low practically. The required PD control of robot joints is well understood and straightforward to implement.

\section{Related Work}

\subsection{Perceptive Locomotion}

To augment traversability of various unstructured terrains, perceptive locomotion offers advantages of proactive terrain negotiation without requiring physical contact or trial and error as in blind walking \cite{miki_learning_2022}. For visual perception, camera feeds are commonly used as input to a neural network to predict safe footholds \cite{Bazeille2013, Siravuru2017}, and LiDAR sensors can be used with proprioceptive information to produce environment-aware control \cite{Fallon2018, Liu2021}. Hybrid methods of RL and trajectory optimization have been used to produce adaptive locomotion from estimated terrain heightmaps \cite{Gangapurwala_2022}. Through a training curriculum, it was shown that motor policies can be learned using sparse visual observations to achieve terrain-aware locomotion \cite{fernando2021}. 

Here, we are motivated to explore the direct use of off-the-shelf onboard cameras (e.g., RealSense D405) without needing complex SLAM information as part of the feedback. We have investigated effective learning-based control framework to use only the front-facing heightmap to encode terrain information and achieve active traversal of various terrains at different gait paces.

\subsection{Motion Imitation for Desirable Behaviors}

DeepMimic \cite{peng_amp_2021} and the related works \cite{peng_learning_2020,smith_finetuning_2021} used motion-capture to generate full and highly-detailed joint trajectories, body position and velocity, body orientation and angular velocity, among others. We refer to such detailed use of reference as \textit{full configuration-level} of motion imitation. The policy is trained with an off-the-shelf RL algorithm, with a reward component to incentivize imitation of the reference trajectories \cite{peng_amp_2021}. 

Although motion imitation incorporates prior knowledge about preferred skills, allowing policy-learning to succeed with much simpler reward functions. However, this too has its own limitation of requiring a suitable dataset of reference motions, which is nontrivial to obtain. Furthermore, the same inductive bias may in turn reduce the flexibility for the policy to learn suitable terrain-adaptive locomotion skills. For example, DeepMimic demonstrated limited ability to navigate ground terrain obstacles as compared to recent methods using vanilla RL \cite{kumar_rapid_2021, miki_learning_2022, Gangapurwala_2022}. 

\subsection{Controllable Gaits using Central Pattern Generators}

Central pattern generators (CPGs) are used here to achieve high-level motion imitation. CPGs have been studied as a simple and effective biomimetic mechanism for locomotion \cite{IJSPEERT_cpg_2008}. Past use of CPGs for locomotion largely focus on the direct generation of a fixed motion pattern in the configuration space \cite{cpg_walking_1, cpg_walking_2, cpg_walking_3, drotman_cpg_pneumatic_legged_2021}. The resulted motion has limited flexibility to adapt to diverse terrains. 

Improving on this, we propose to use CPGs to generate \textit{foot-ground references} to \textit{guide} a learning policy, such that it is incentivized to imitate the reference gait and also has more flexibility to optimize around the \textit{foot-ground references} to traverse terrain obstacles more successfully. 
It is clear that our CPG-guided framework is inherently different from the traditional way of combining CPG joint angle trajectories with neural network residual joint offsets. Our proposed method has the advantage that the learning has more flexibility to learn richer terrain-adaptive motions in the joint space.  

Recently, \cite{shao_learning_2022} demonstrated the advantages and feasbility of CPGs in generating reference foot position trajectories, which were subsequently used as a motion imitation primitive in combination with a learning-based locomotion framework. However, the analysis was limited to a small number of discrete, fixed gaits. We extend this further to demonstrate feasibility across a continuous range of gait parameters. Furthermore, we reduce the bias towards a particular fixed locomotion behavior by replacing the reference foot position trajectories with reference \textit{foot contact} states. Our new proposal allows the policy more leeway to explore and learn terrain-adaptive locomotion. 

\section{Methods}

\subsection{Proposed Architecture of Perceptive Locomotion}

\begin{figure*}[h]
    \centering
    \includegraphics[width=\linewidth]{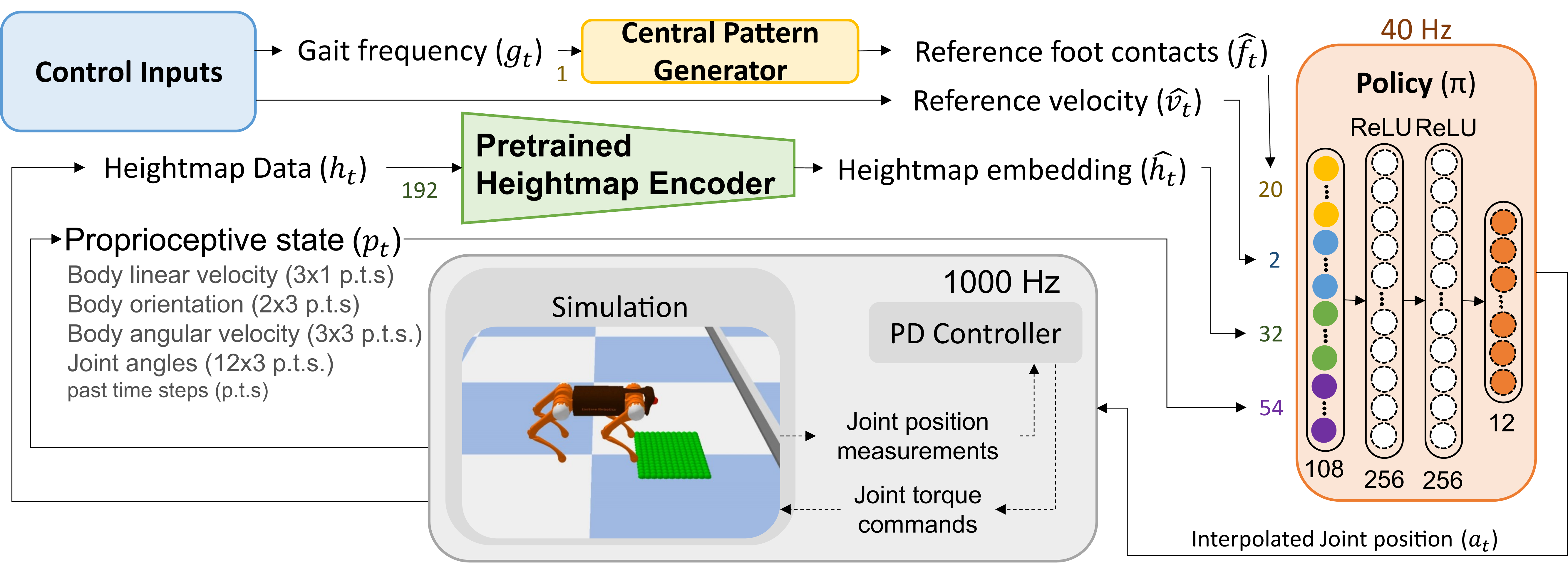}
    \caption{Architecture of learning perceptive locomotion by encoding both visual and proprioceptive feedback, with CPG-based gait regulation. Additional task-control inputs and CPG-based foot-ground contact regulation allow to adjust trotting velocity and gait frequencies on-the-fly. Terrain perception is represented by a latent vector from a heightmap encoder pretrained via the scanning of various terrain surfaces.}
    \vspace{-1mm}
    \label{fig:architecture}
\end{figure*}

The control architecture for our learning-based control policy is presented in Fig. \ref{fig:architecture}. The architecture is identical during training and testing. At every time step $t$, the policy receives a state observation $s_t = (p_t, \hat{h_t}, \hat{f_t}, \hat{v_t})$ containing proprioceptive state vector $p_t$, heightmap embedding $\hat{h_t}$, reference foot contacts $\hat{f_t}$, and reference velocity $\hat{v_t}$. The proprioceptive state $p_t \in \mathbb{R}^{54}$ includes the agent's linear velocity $(v^t_x, v^t_y, v^t_z)$, and a 3-step history of body orientation $(\theta^{t-2:t}_{roll}, \theta^{t-2:t}_{pitch})$, body angular velocity $(\omega^{t-2:t}_{roll}, \omega^{t-2:t}_{pitch}, \omega^t_{yaw})$, and joint angles $(q^{t-2:t})$.

The heightmap embedding $\hat{h_t} \in \mathbb{R}^{32}$ consists of the encoded output of the distance measurements obtained from the $12 \times 16 = 192$ rays used in the heightmap setup. The heightmap measurement provides distance to physical obstacle, and is clipped between 0.1 m and 8.0 m to replicate the specifications of commercially available RGB-D cameras. More details on the heightmap setup and encoder can be found in Section \ref{sec:visual}.

The reference foot contacts $\hat{f_t} \in \mathbb{R}^{20}$ includes $(\hat{f_{t}}, \hat{f_{t+1}}, \hat{f_{t+2}}, \hat{f_{t+10}}, \hat{f_{t+50}})$. More details on the reference foot contacts and central pattern generator can be found in Section \ref{sec:cpg}. From our experiments, allowing the policy to observe future reference states enables better tracking of the reference foot contact trajectory. In this paper, the indices are chosen to achieve a balance between short-term and long-term future observations.

The target velocity vector $\hat{v_t} \in \mathbb{R}^2$ specifies the speed at which the agent should move along the $x$ and $y$ axes. The policy $\pi$ outputs action $a_t$, which are the desired joint positions. These are fed into joint PD control to produce joint torques for simulating dynamics in the physics engine. The corresponding state measurements are then fed back into the policy to close the feedback loop.

\subsection{Learning Locomotion via Reinforcement Learning}

We formulate perceptive locomotion as an RL problem. Given states $s_t$ and Markovian dynamics $p(s_{t+1}, r_t | s_t, a_t)$ where $a_t$ are actions and $r_t$ is the reward, the objective is to learn a control policy $\pi(a_t | s_t)$ that enables an agent to maximize its expected return for a locomotion task. 

At each time step $t$, the agent observes a set of states $s_t$ from the environment and samples a set of actions $a_t \sim \pi(a | s_t)$ from its policy $\pi$. The agent then applies $a_t$, which results in a new state $s_{t+1}$ and a scalar reward $r_t = r(s_t, a_t, s_{t+1})$. Such a repeated process generates a trajectory $\sigma = \{(s_k, a_k, r_k)\}_{t \in \mathbb{N}}$. The objective is to maximize the expected return $J(\pi)$: 
\begin{equation}
    J(\pi) = \mathbb{E}_{\sigma \sim p(\sigma | \pi)} \bigg[\sum_{t=0}^{N-1} \gamma^t r_t\bigg ]
\end{equation}
where $N$ denotes the duration of the episode, and $\gamma \in [0,1]$ a time-discount factor. Given that $r_t$ is bounded, the sum on the R.H.S converges to a finite limit. $p(\sigma | \pi)$ denotes the likelihood of a trajectory $\sigma$ under policy $\pi$:
\begin{equation}
    p(\sigma|\pi) = p(s_0)\prod_{t=0}^{N-1} p(s_{t+1}|s_t, a_t)\pi(a_t | s_t)
\end{equation}
with $p(s_0)$ being the initial state distribution, and $p(s_{t+1}|s_t, a_t)$ representing the dynamics of the system, which determines the effects of the agent's actions.

\subsubsection{Reward Design for Perceptive Locomotion}

The reward function encourages the robot to move at the given reference velocity $\hat{v_t} \in \mathbb{R}^2$ and follow sequences of reference foot contacts, and penalizes it for unstable and inefficient motions. At every time step $t$, we denote the agent's linear velocity as $\boldsymbol{v}_t$, its four foot contacts as vector $\boldsymbol{f}_t$, orientation as $\boldsymbol{\theta}_t$, joint torques as $\boldsymbol{\tau}_t$, joint velocities as $\boldsymbol{\dot{q}}_t$.

The reward term to encourage following the reference velocity $\hat{v_t}$ is given by:
\begin{equation}
    r^t_v =
    \begin{cases}
    e^{-c_{\alpha}(d_{tgt})^2} & \text{if $d_{tgt}<0$} \\
    e^{1-\frac{1}{1-c_{\beta}(d_{tgt})^2}} & \text{if $d_{tgt}\geq0$ and $d_{tgt}<\frac{1}{c_{\beta}}$} \\
    0 & \text{otherwise}
    \end{cases}
\end{equation}
where $d_{tgt} = \frac{\hat{\boldsymbol{v}_t} \cdot \boldsymbol{v}_t}{||\hat{\boldsymbol{v}_t}||^2}$ is the normalized distance from base and target velocity vectors, and $c_{\alpha}$ and $c_{\beta}$ are pre-determined reward parameters as $c_{\alpha}=16.0$ and $c_{\beta}=25.0$ for training the policy.

A Gaussian function is used when the agent's velocity is less than the target velocity in the target velocity's direction. This is to help the agent to learn with less restricted penalty that allows it to slow down and engage with the terrain when necessary. On the other hand, a bump function is used when the agent's velocity is more than the desired one in the targeted direction. This is to enforce the policy to follow the given reference velocity.

The reward term to penalize instability is given by:
$
    r^t_s = -||
    \begin{bmatrix}
        0, 0, 0 \\
        0, 0, 0 \\
        1, 1, 0 \\
    \end{bmatrix}
    \cdot \boldsymbol{R}_t||
$
, where $\boldsymbol{R}_t$ is the rotation matrix derived from the quaternion of the agent's orientation $\boldsymbol{\theta}_t$.
The reward term to encourage following the reference foot contacts from control input is given by:
$r^t_f = \frac{1}{4}(\hat{\boldsymbol{f}_{t, 0}} \cdot \boldsymbol{f}_t)$.
The reward term to penalize energy usage is given by:
$
    r^t_e = -|\boldsymbol{\tau}_t \cdot \boldsymbol{\dot{q}}_t|
$.
For training the policy, the total reward at any time $t$ is defined as the sum of the weighted reward terms:
$
\label{eq:reward}
r_t = w_v * r^t_v + w_f * r^t_f + w_s * r^t_s + w_e * r^t_e
$, where $w_v = 1.0$, $w_f = 2.0$, $w_s = 1.5$, $w_e = 0.0001$.

\subsubsection{Training Curriculum}

This work focuses on quadruped trotting as it is the most common and useful gait in real-world applications. The gait frequency is sampled uniformly within $[1.5, 2.5]$ Hz, and the reference velocity is sampled uniformly within $[0.4, 0.8]$ m/s. Gait frequency and target velocity are sampled at the start of each episode and are held constant throughout the episode. 

Three platforms and stairs of different heights are used in training - 2 cm, 5 cm, and 7 cm. We use a fixed-order curriculum for the terrains during training: 2 cm $\rightarrow{}$level 1; 5cm $\rightarrow{}$level 2; 7 cm $\rightarrow{}$level 3. At the start of the training, the agent begins at level 1. When the agent has successfully traversed up and down the terrain at the current level, then its level is increased by one. Then at the next reset, the agent will start on a randomly selected terrain at its new level or lower. The next selected terrain is proportionally sampled according to its difficulty.

\subsubsection{Terrain Descriptions}
Other than (i) \textit{platforms}, we also tested the trained neural network controller on other terrains:
(ii) \textit{Hurdles}: narrow obstacles designed to evade perception and trip the robot. They have a step run of 0.06 m. In Figures \ref{fig:app-gait-transition} and \ref{fig:app-robustness}, the step height is specified between 0.03 and 0.06 m.
(iii) \textit{Heightfield}: a rugged terrain with obstacles protruding from the ground. The specified heightfield terrain area is split into 0.1 m $\times$ 0.1 m grids, where each grid unit is set to a random height uniformly selected between 0 and the specified perturbation height. In Figures \ref{fig:app-gait-transition} and \ref{fig:app-robustness}, the perturbation height is set to 0.05 m.
(iv) \textit{Stairs}: staircases with 5 steps. Each step has a run of 0.3 m. In Figures \ref{fig:app-gait-transition} and \ref{fig:app-robustness}, the range of the height of steps is specified between 0.03 and 0.07 m.

\subsection{Exteroceptive Perception of Terrains}
\label{sec:visual}

\begin{figure}[t]
    \centering
    \includegraphics[width=0.9\linewidth]{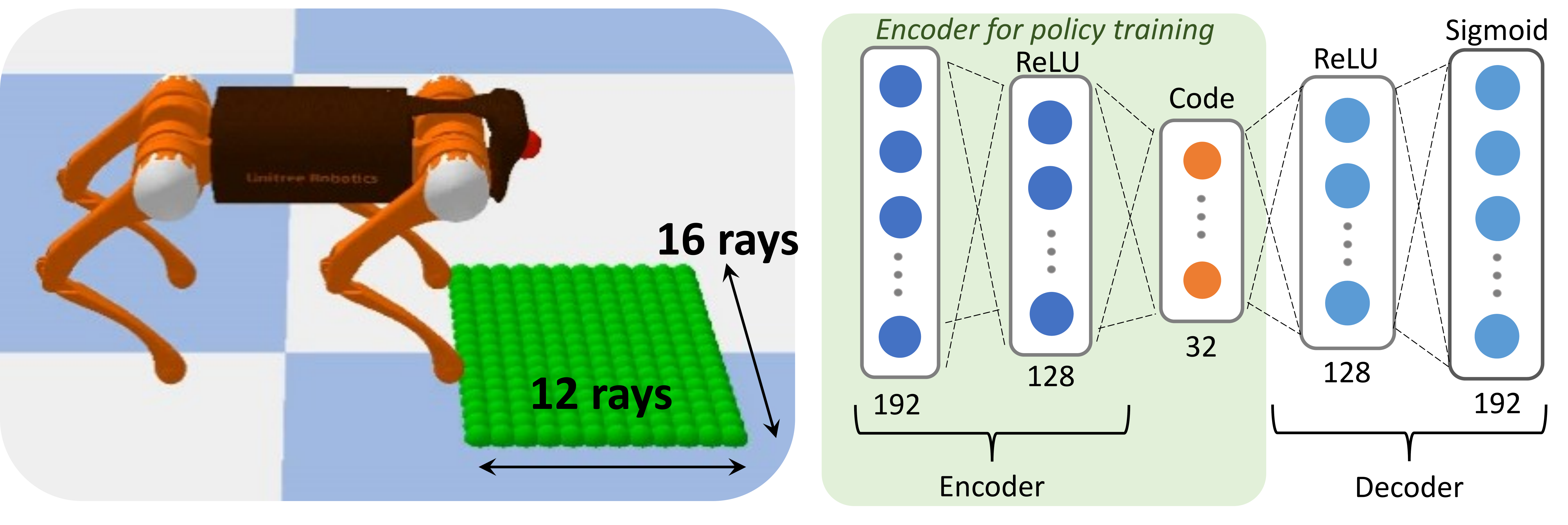}
    \caption{Configuration of visual perception and its feature encoding: (a) Setup of heightmap; (b) Heightmap autoencoder, and the green shaded area is the encoder used in policy training.}
    \vspace{-2mm}
    \label{fig:heightmap}
\end{figure}

The robot updates the heightmap data $h_t$ in front of it at each time step $t$. As seen in Fig. \ref{fig:heightmap}, the heightmap is designed with a dimension of 12 $\times$ 16 rays, each grid in the map is spaced horizontally and vertically 0.03 m apart and contains the height coordinate $z$ of the scanned area. With a robot base height of 0.35 m, the field of view of is about 70\textdegree $\times$ 55\textdegree, which can be easily covered by commercial-off-the-shelf RGB-D cameras.

To achieve such dimension reduction, we first train an autoencoder to compress the heightmap data, see Fig. \ref{fig:heightmap} for the overall architecture. Mean square error is used as the loss function. Heightmap data is collected from simulation with a camera fixed on a ``hovering" robot across different terrain types. We used only the encoder part of the trained heightmap autoencoder, and compressed the exteroceptive heightmap observation $h_t \in \mathbb{R}^{12\times16}$ to a lower dimension state of $\hat{h_t} \in \mathbb{R}^{32}$. During the training of motor policy, the weights of all layers in the encoder are fixed to produce heightmap embedding $\hat{h_t}$, which is fed into the policy. 

\subsection{Central Pattern Generators and Gait Representation}
\label{sec:cpg}

Our motivation of using CPGs is to have active control of the gait pattern in the form of foot-ground contact sequences, which gives the RL agent more flexibility to explore joint-space motions and diverse terrain-adaptive skills. Our specifically formulated CPG approach can control both the contact frequency and duty cycle of the trotting gait.

Traditional RL approach normally specifies target velocity instead of contact patterns. Hence, for different velocities, the robot changes the step length only. In contrast, our method adapts from \cite{shao_learning_2022} and formulates CPGs to generate a total of 4 abstract phases (one per leg). The abstract phase vector is decoded to a reference foot contact state, which is used as state input for the neural network. As a result, the CPG does \textit{not} impose joint angles, but guides the policy only in terms of ground contacts. With the same foot-contact references, the foot clearance and stride length are free to explore, enabling a considerable range of terrain-adaptive behaviors. 

\begin{figure}[t]
\vspace{-3mm}
    \centering
    \includegraphics[width=\linewidth]{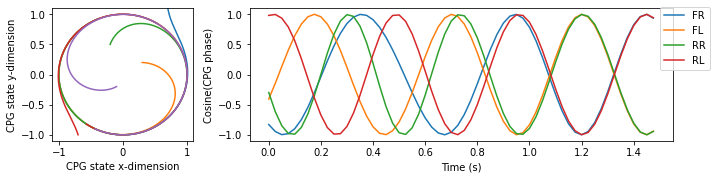}
    \caption{\label{fig:single_and_coupled_cpg}Advantages of CPG: (a) Convergence to a stable limit cycle (Equation \ref{eqn:cpg}); (b) Four synchronized coupled Hopf oscillators for trotting (Equation \ref{eqn:coupled_cpg}).}
    \vspace{-1mm}
\end{figure}

We found that without the CPG input, the quality of learned locomotion degrades significantly. For conciseness, we provide a brief recapitulation below. A CPG is a dynamical system that produces rhythmic patterns with a stable limit cycle. Fig. \ref{fig:single_and_coupled_cpg} shows the Hopf oscillator as an example of a CPG, with converging limit cycles. A Hopf oscillator has parameters $(\alpha, \beta, b, T, \mu, \gamma)$ and state $\rho= (x,y) \in \mathbb{R}^2$, radius $r = \sqrt{x^2 + y^2}$ with intrinsic update equations: 
\begin{equation}
\label{eqn:cpg}
    \dot x = \alpha(\mu^2 - r^2)x + \gamma y, \quad \dot y = \alpha(\mu^2 - r^2)y - \gamma x
\end{equation}
In our work, parameters are $\alpha = 10.0, b = 50.0, \mu = 1.0, \gamma = 50.0$. $T, b$ are the parameters to modulate the desired gait frequency and duty factor respectively.

We initialize four Hopf oscillators with identical parameters, one for each leg. It is desired that the four oscillators are synchronized, maintaining a fixed \textit{phase offset} $\phi \in \mathbb{R}^4$, where $\phi_i$ is the desired phase of CPG $i$. To enable synchronization of the four oscillators, we modify the intrinsic update equations in (\ref{eqn:cpg}) with a coupling term $R_{ij}$: 
\begin{equation}\label{eqn:coupled_cpg}
\begin{aligned}
    & \dot \rho_i = f(\rho_i) + \sum_{ij} R_{ij}\rho_j, \quad R_{ij} = \begin{bmatrix}
        \cos \theta_{ij} & - \sin \theta_{ij} \\ 
        \sin \theta_{ij} & \cos \theta_{ij} \\
    \end{bmatrix} \\ 
    & \theta_{ij} = \phi_j - \phi_i, \quad i,j \in \{1,2,3,4\}
\end{aligned}
\end{equation}
Fig. \ref{fig:single_and_coupled_cpg} shows an example of how four oscillators beginning from a random initial state can achieve convergence when they are updated according to Equation \ref{eqn:coupled_cpg}.

Gaits are parameterized as binary reference trajectories $f \in \{-1,1\}^{N \times 4}$. To generate $f_{ti}$, where $t$ is time and $i$ is the leg index, for each oscillator state $(x_{ti}, y_{ti})$, the corresponding leg phase is $\hat\theta_{ti} = \arctan(\frac{y_{ti}}{x_{ti}})$, which is decoded to a binary foot contact state as $f_{ti} = 2 \cdot \mathrm{I}[\hat\theta_{i} > 0] - 1$.

\begin{figure}[t]
    \centering
    \includegraphics[width=60mm]{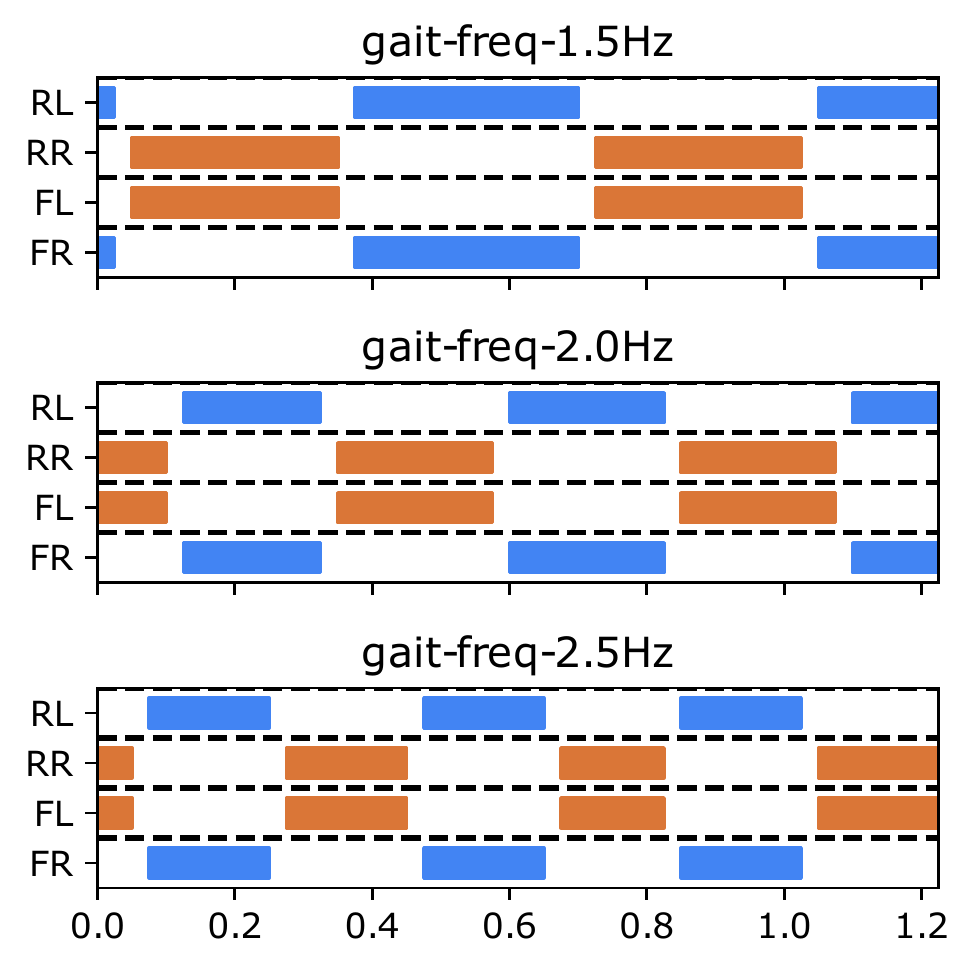}
    \caption{Foot-contact references with 0.5 duty factor and varying trotting frequencies.}
    \label{fig:gait_repr}
     \vspace{-4mm}
\end{figure}

As a result, we obtain $f$ where $f_{ti} = 1$ if and only if foot $i$ should be on the ground at time $t$, and $-1$ otherwise. This method of representing gaits as binary trajectories enables concise encapsulation of key properties: (i) gait \textit{frequency} which controls how rapidly a rhythmic gait is generated, (ii) gait \textit{duty factor} which represents the stance-to-swing duration ratio, and (iii) \textit{phase offsets} which determines the class of gait being executed (e.g. walk, trot, canter, pace). In this work, we focus on the trotting gait which is the mostly used in real applications, and therefore only regulate the frequency part and keep the latter two constant. Fig. \ref{fig:gait_repr} shows the trajectories of binary foot-ground contract for each gait frequency.

Our CPG-guided framework provides a balance between vanilla RL and direct motion imitation. By incorporating an imitation term of contact reference, we create an inductive bias towards known stable gait patterns, while keeping the simplicity of reward formulation. Because the desired joint motions are not hard-coded in the contact references, there is considerable flexibility for the RL policy to learn terrain-adaptive skills. 


\section{Simulation Results}

\subsection{Setup of Simulation and Training}

The NN policy is an actor-critic architecture with parameter sharing. The shared backbone is a 2-layer network, each having 256 nodes and ReLU activation. The linear prediction heads in the last layer are used to obtain the value estimate and action distribution. 

The NN policy is queried at a control frequency of 40 Hz, which predicts desired local joint angles $a = \hat{\boldsymbol{q}} \in \mathbb{R}^{12}$. The target joint positions are post-processed as: (i) clipped to a maximum of 0.5 radians from the default resting pose; (ii) velocity is limited by a maximum change of 0.15 radians from the current position; and (ii) filtered using a first-order Butterworth low-pass filter in case of outliers or spikes. Using PD control, such updated actions $\hat{a}$ generate joint torques $\tau$ as: $\tau = K_p (\hat{\boldsymbol{q}} - \boldsymbol{q}) + K_d (\hat{\dot{\boldsymbol{q}}} - \dot{\boldsymbol{q}})$. $K_p$ and $K_d$ are the proportional and derivative gains, and the target joint velocities $\hat{\dot{\boldsymbol{q}}}$ are 0 here. For each leg, $K_p$ are $(100, 100, 100)$ and $K_d$ are $(1, 2, 2)$ for the hip, thigh, and knee joints respectively.

\begin{wraptable}{r}{0.45\linewidth}
\vspace{-6mm}
\small
\setlength{\tabcolsep}{3pt}
\begin{tabularx}{\linewidth}{|l|X|}
    \hline
    {Parameter}  & {Value} \\
    \hline
    {Discount factor $\gamma$} & {0.95} \\
    \hline
    {Learning rate} & {$3\mathrm{e}-5$} \\
    \hline
    {PPO clip threshold} & {0.4} \\
    \hline
    {PPO batch size} & {1024} \\
    \hline
    {PPO epochs} & {20} \\
    \hline
\end{tabularx}
\caption{PPO hyperparameters for training.}
\label{tab:hyperparams}
\vspace{-1mm}
\end{wraptable}

The learning task is implemented as an OpenAI Gym \cite{openai_gym} environment and uses PyBullet \cite{coumans2021} for the physics, rigid-body and contact dynamics simulated at 1000 Hz. At each simulation time step, the latest target joint position is used in PD control to generate motor torques. This work used the simulated Unitree A1 robot \cite{unitree} and its associated kinematic and dynamic parameters.

The policy was trained using Proximal Policy Optimization (PPO) \cite{ppo} implemented in Stable Baselines3 \cite{stable-baselines3}. See Table \ref{tab:hyperparams} for the hyperparameters used. Each training episode lasts for a maximum of 1000 discrete time steps. To reduce unnecessary computation, we employ early termination of episodes when the robot loses balance.

The learned policy was evaluated on the following terrains: (i) \textit{Platforms}, raised platforms of progressively increasing height; (ii) \textit{Hurdle}, narrow obstacles designed to evade perception and trip the robot; (iii) \textit{Heightfield}, rugged terrain with obstacles protruding from the ground; and  (iv) \textit{Stairs}, staircases of progressively increasing height.

\subsection{Controllable Trotting Gait Frequencies and Speed}

\begin{figure}[t] \hfill
  \centering
  \includegraphics[width=\linewidth]{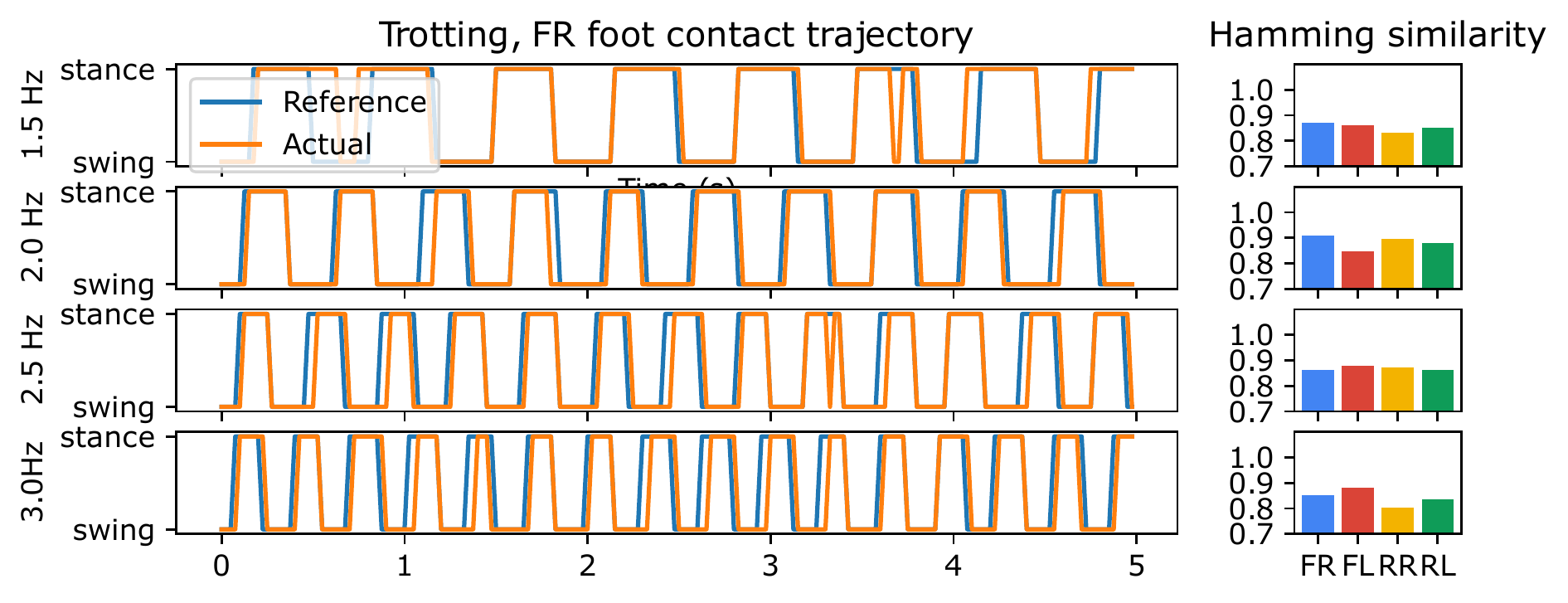}
  \vspace{-7mm}
  \caption{Contact phases of the front right foot across different reference gait frequencies at reference velocity 0.6 m/s, and Hamming similarity across all feet.
  \vspace{-4mm}
  \label{fig:gait_imitation}}
\end{figure}

The trained policy is designed to be able to regulate contact phases, and thus the frequency of the trotting pace. Fig. \ref{fig:gait_imitation} shows the executed foot contacts closely followed different reference gait frequencies of 1.5, 2.0, and 2.5 Hz, at a fixed target velocity of 0.6 m/s. A Hamming similarity of approximately 0.8 is achieved across all feet contacts.

\begin{figure}[t]
    \centering
    \includegraphics[width=1.0\linewidth]{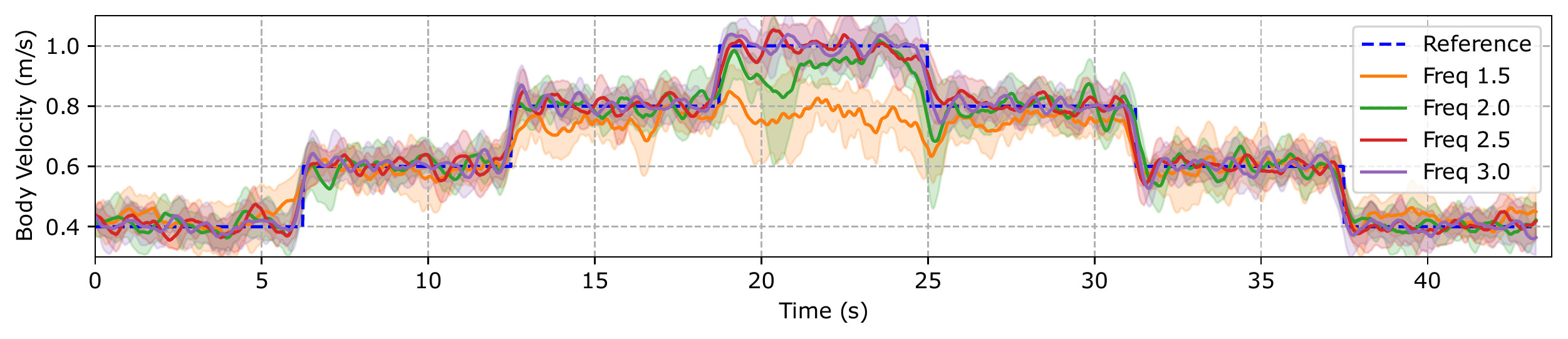}
    \vspace{-7mm}
    \caption{Traveling velocities measured across tests at different \textit{reference} gait frequencies.}
    \vspace{-7mm}
    \label{fig:results-freq-vel}
\end{figure}

At different gait frequencies, the trained policy can simultaneously follow a given reference velocity. Fig. \ref{fig:results-freq-vel} shows the body velocities measured during 4 sets of simulation tests with gait frequencies of 1.5, 2.0, 2.5 and 3.0 Hz. Each set of tests used reference forward velocities of 0.4, 0.6, 0.8, and 1.0 m/s. A total of 20 trials with different random generator seeds were run per reference gait frequency.
The systematic testing shows that a too low-frequency gait of 1.5 Hz is unable to catch up with a comparatively high velocities of 0.8 to 1.0 m/s, given the leg length of the robot. Needless to say, higher-frequency gaits of 2.0, 2.5 and 3.0 Hz are able to do so. Results in Fig. \ref{fig:results-freq-vel} show that with higher gait frequency, meaning a faster pace, the robot follows 1.0 m/s reference better. This implies a natural correlation between gait frequency and achievable traveling speed, given the size of a physical robot, i.e. the higher the gait frequencies, the faster the robot can trot.

\subsection{Terrain-Adaptive Gait with Natural Transitions}

Further validations were done to demonstrate that the learned policy can achieve seamless transitions between gaits. This allows to select gait parameters and enforce how fast the robot navigates over different obstacles, e.g., cautious slow pace on difficult surfaces and faster gait on flat ground for maximizing task efficiency.

\begin{figure}[t]
\centering
\vspace{-2mm}
    \includegraphics[width=1.0\linewidth]{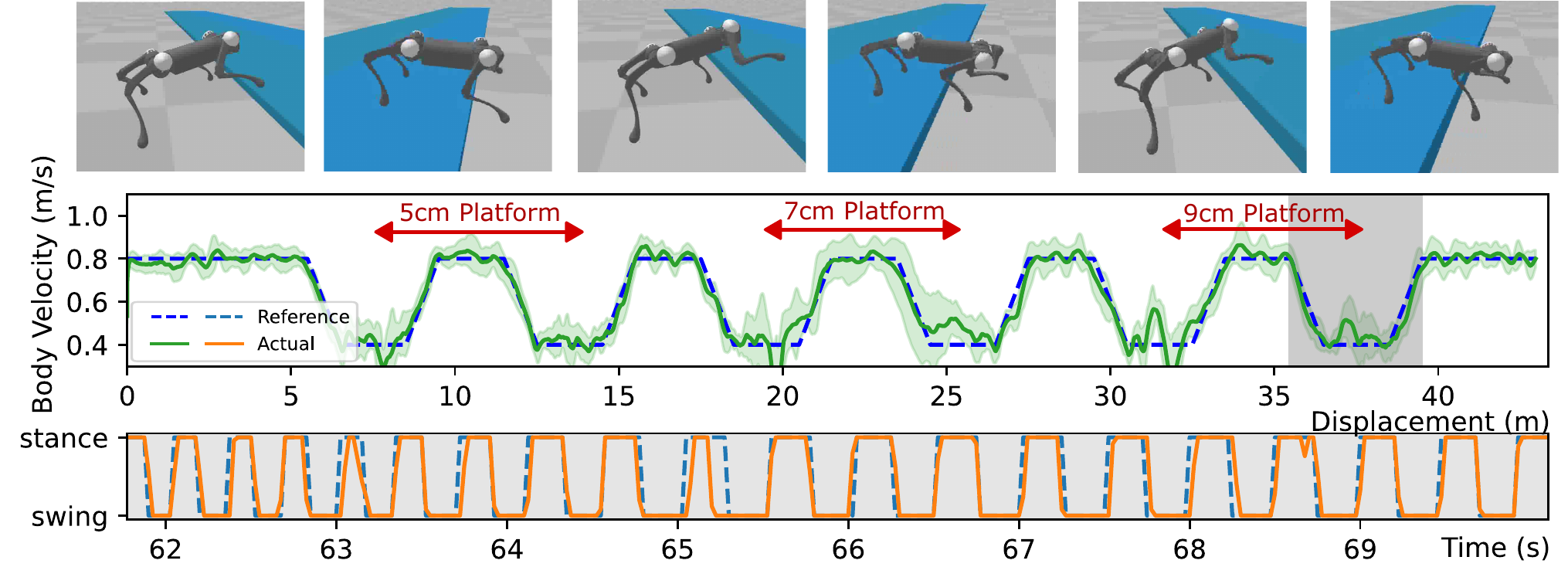}
    \includegraphics[width=1.0\linewidth]{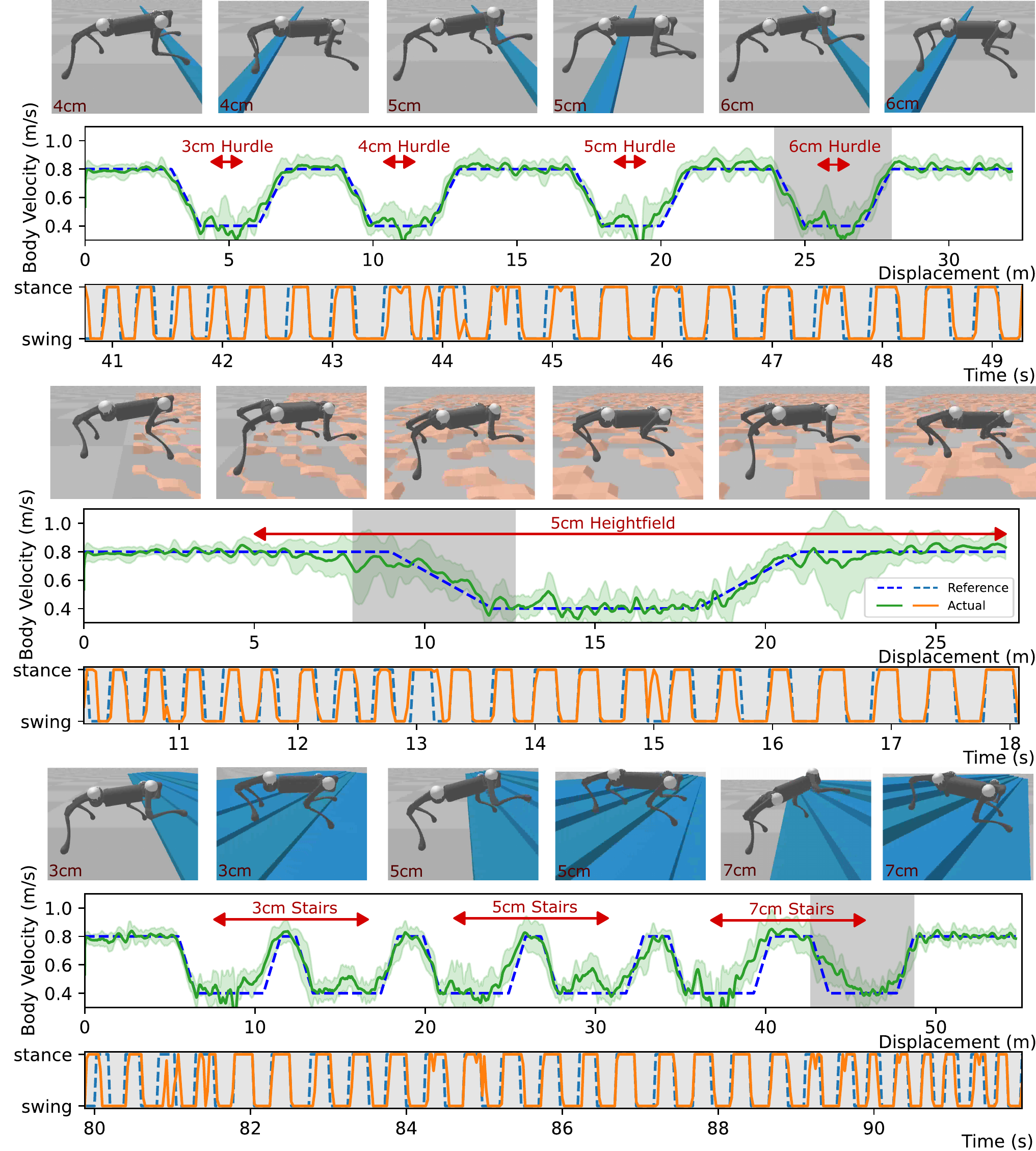}
    \vspace{-5mm}
    \caption{Measured body velocities and foot contacts which were tested across different terrains with controlled gait transitions. The reference gait frequency and velocity start to decrease while encountering an area with rough terrains or obstacles.}
    \vspace{-3mm}
    \label{fig:results-gait-transition}
\end{figure}

We evaluated such capability of terrain-adaptive gait transitions by using scheduled control inputs (see Fig. \ref{fig:architecture}), which actively modulate gait transitions as the robot is crossing terrain obstacles. On flat ground, a gait of 3.0 Hz and 0.8 m/s is used. When encountering terrain obstacles, this is linearly reduced to 0.4 m/s. A total of 20 trials were tested per terrain. Fig. \ref{fig:results-gait-transition} shows that the robot can smoothly follow the varying control inputs while crossing various terrain. 

\begin{figure}[!t]
\centering
\includegraphics[width=1.0\linewidth]{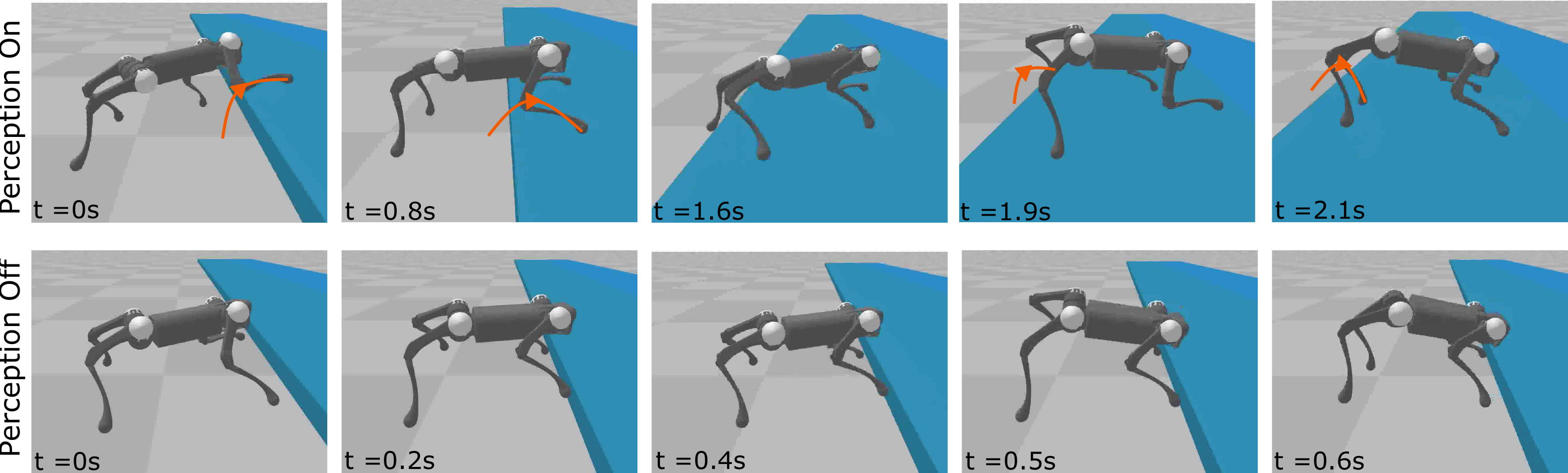}
\caption{Comparison of locomotion over a 9cm platform with and without visual perception.}
\vspace{-5mm}
\label{fig:results-perception}
\end{figure}

\begin{figure}[t]
    \centering
    \includegraphics[width=1.0\linewidth]{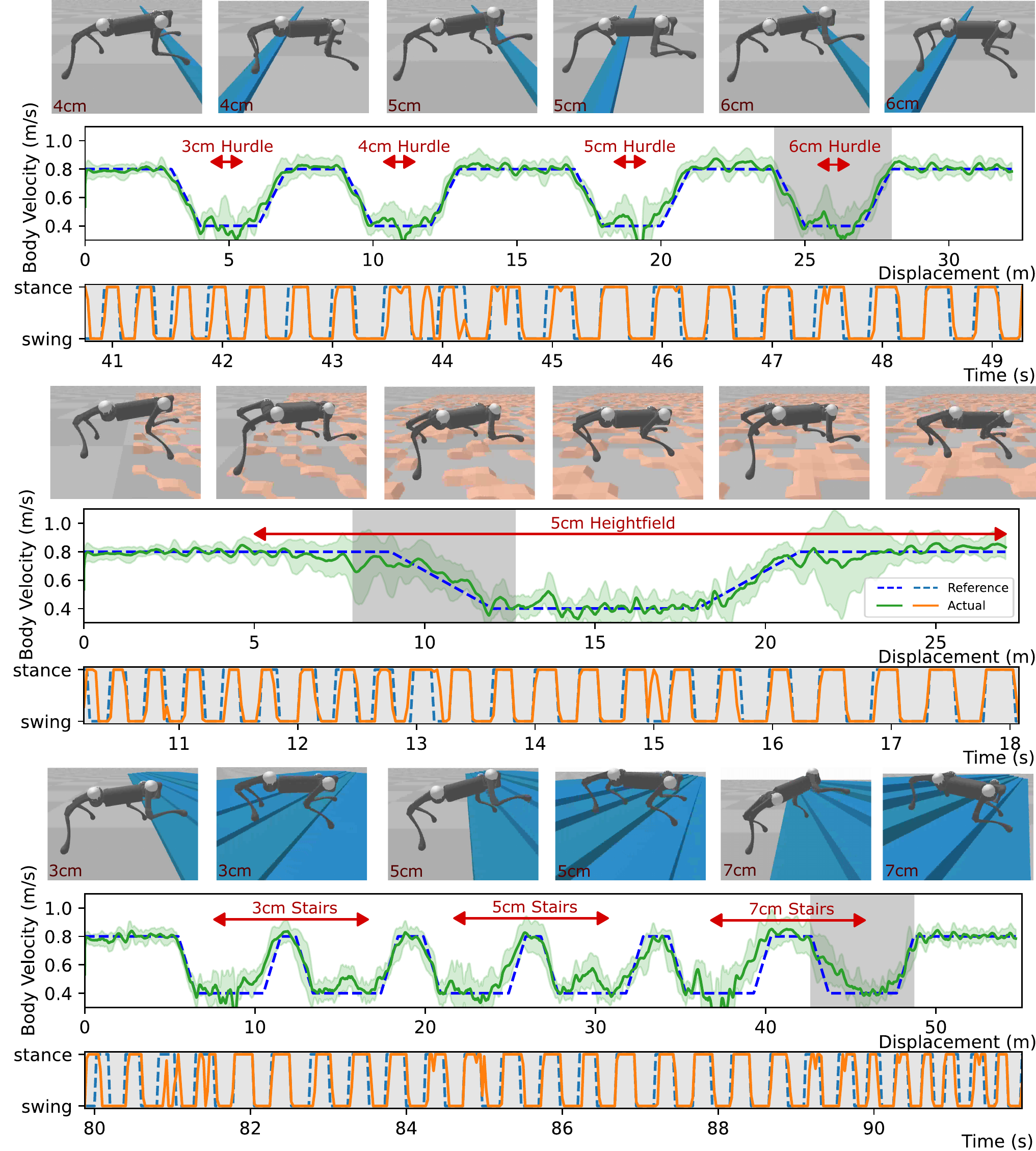}
    \caption{Measured body velocities and foot contacts which were tested across different terrains with controlled gait transitions. The reference gait frequency and velocity start to decrease while encountering an area with rough terrains and/or obstacles.}
    \label{fig:app-gait-transition}
\end{figure}

Figure \ref{fig:app-gait-transition} shows that the robot can smoothly follow the varying control inputs while crossing other terrains than just platforms. On flat ground, a gait of 3.0 Hz and 0.8 m/s is used. When encountering terrain obstacles, this is linearly reduced to 2.0 Hz and 0.4 m/s. A total of 20 trials with different random generator seeds were tested per terrain.

\subsection{Robustness and Viability Tests}

Due to unavailability of hardware, we tested the robustness in simulation to see viability to implement on real robots. All our tests have validated that the physical constraints, i.e., joint positions, velocities, and motor torques are always within the hardware limits. 

Fig. \ref{fig:results-perception} demonstrates the effectiveness of our designed perception in the traversal over uneven terrains. By perceiving the heightmap, the robot is able to \textit{proactively} step over a 9~cm platform. However, without perception, the 9~cm-step is comparatively high (body height is 25~cm during trotting); relying on the proprioceptive data only, the robot failed to achieve sufficient foot-ground clearance blindly and is hindered from moving forward.

\begin{figure}[t]
    \centering
    \includegraphics[width=1.0\linewidth]{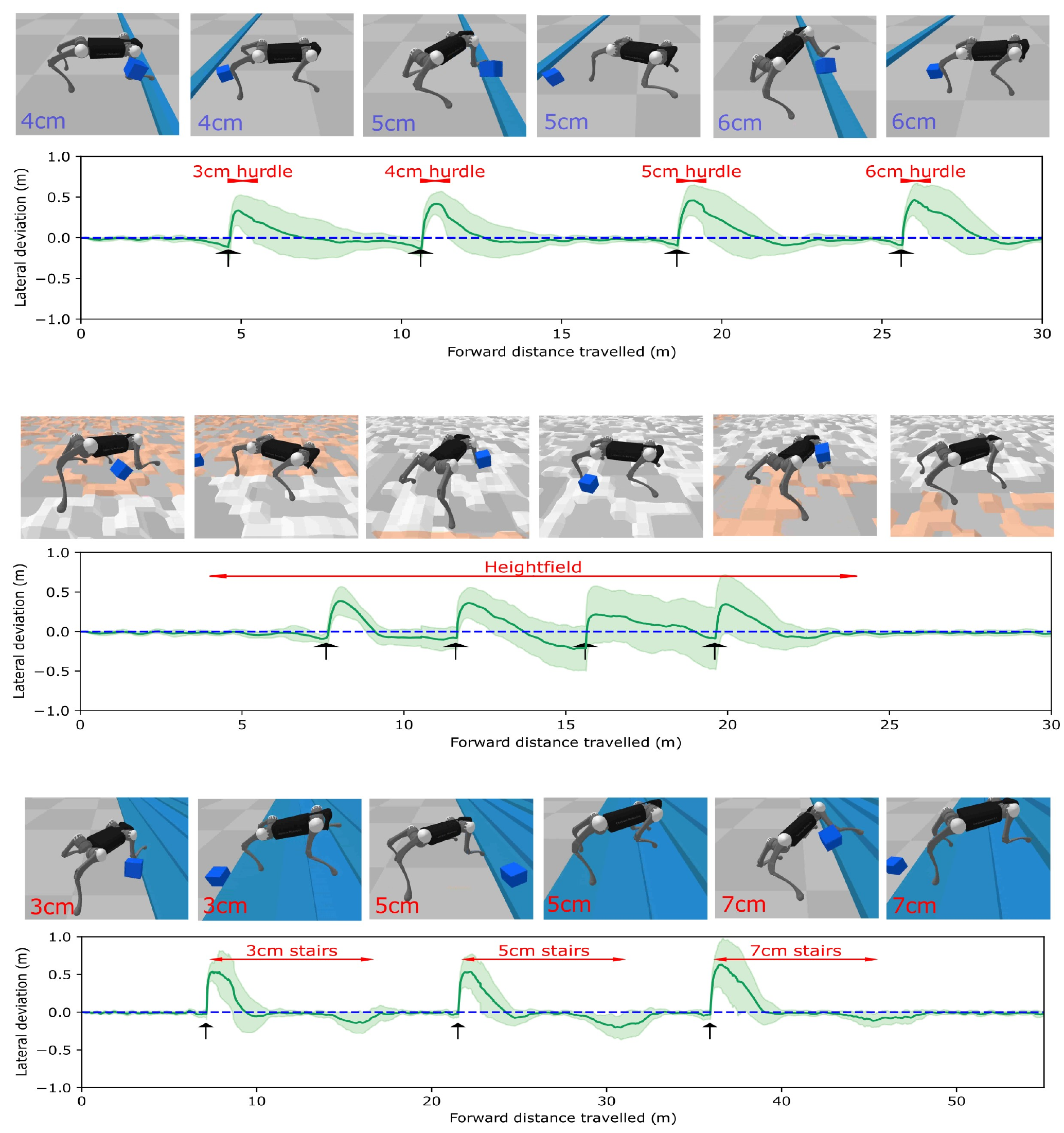}
    \caption{Robust perceptual locomotion over various terrains by rejecting impact disturbances. Black arrows indicate the timing and direction of the flying objects. In all test cases, the learned policy robustly recovers back to the reference path.}
    \label{fig:app-robustness}
\end{figure}

Figure \ref{fig:app-robustness} shows that the robot can robustly recover from unseen collisions and return to the reference path while crossing other terrains than just platforms. On flat ground, a gait of 3.0 Hz and 0.8 m/s is used. When encountering terrain obstacles, this is linearly reduced to 2.0 Hz and 0.4 m/s, similar to Figure \ref{fig:app-gait-transition}. A total of 20 trials with different random generator seeds were tested per terrain.

\begin{figure}[!t]
\centering
\includegraphics[width=0.15\linewidth]{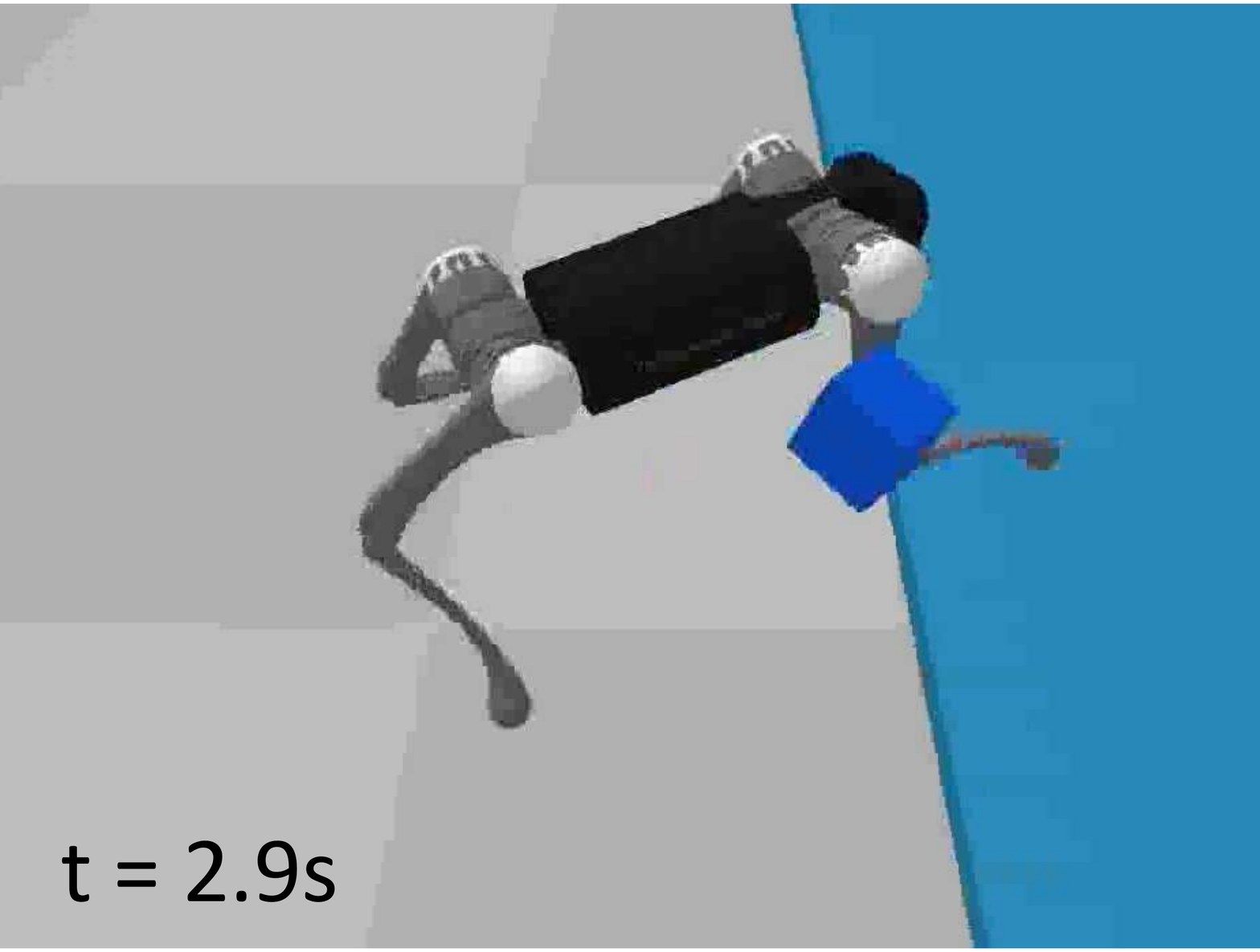}
\includegraphics[width=0.15\linewidth]{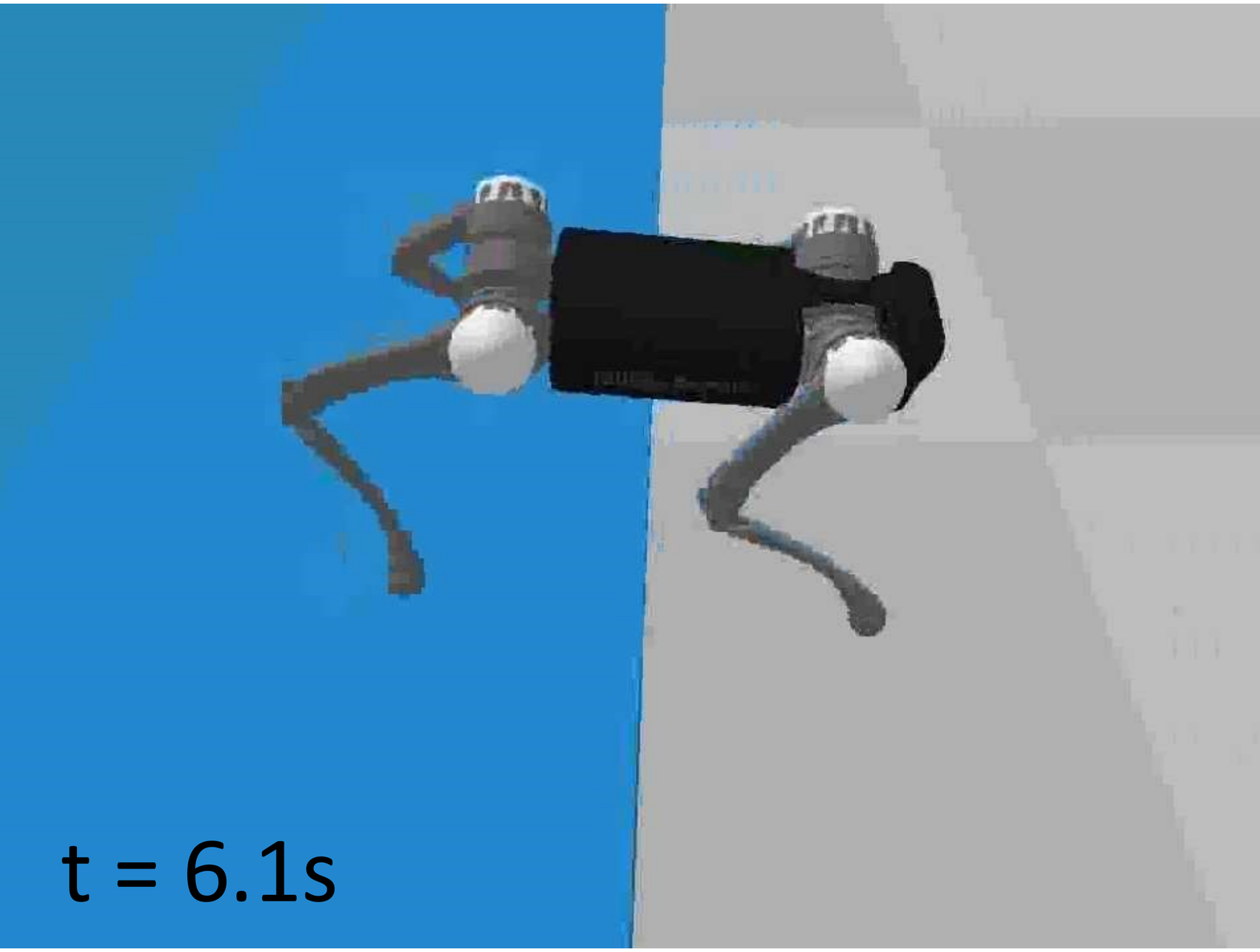}
\includegraphics[width=0.15\linewidth]{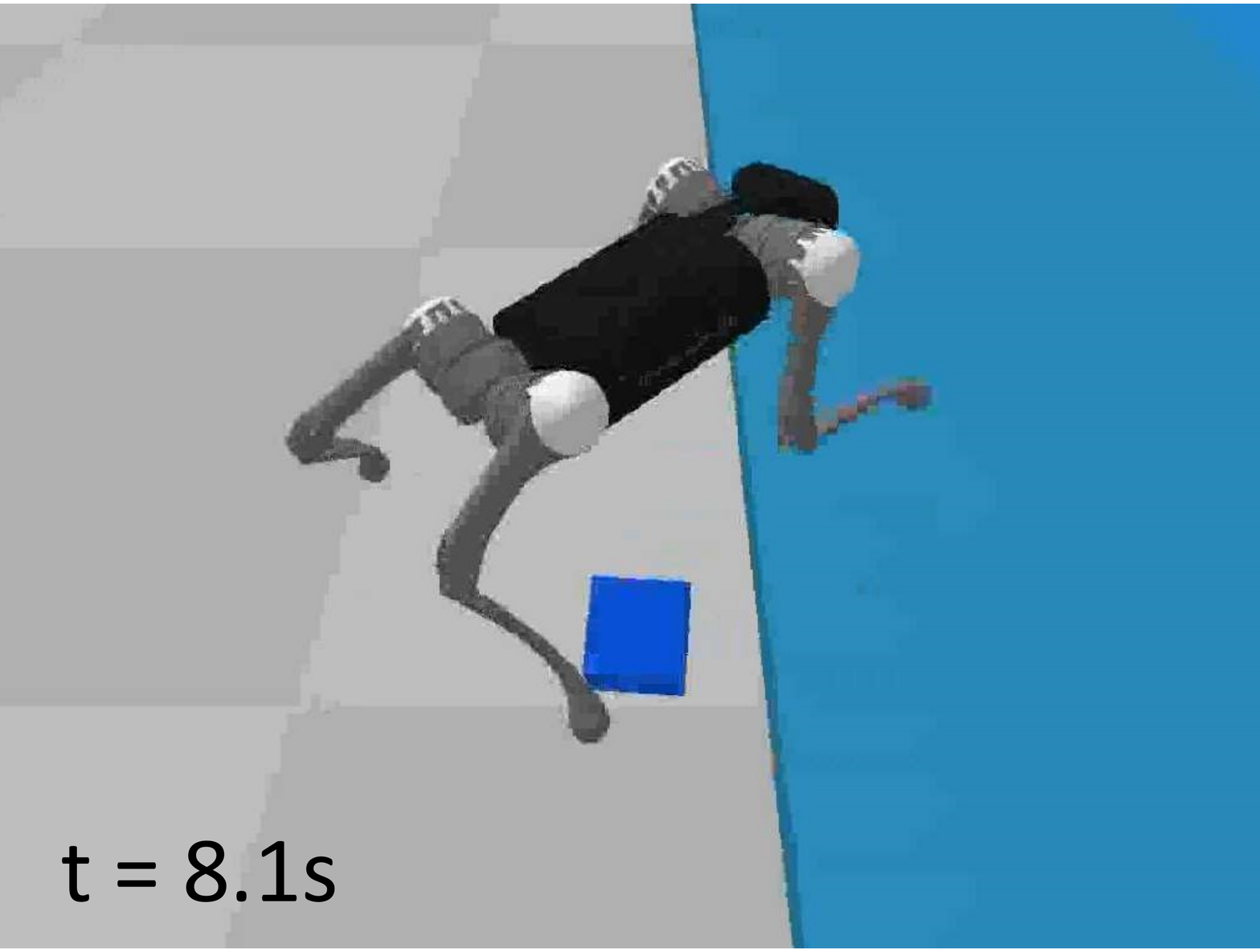}
\includegraphics[width=0.15\linewidth]{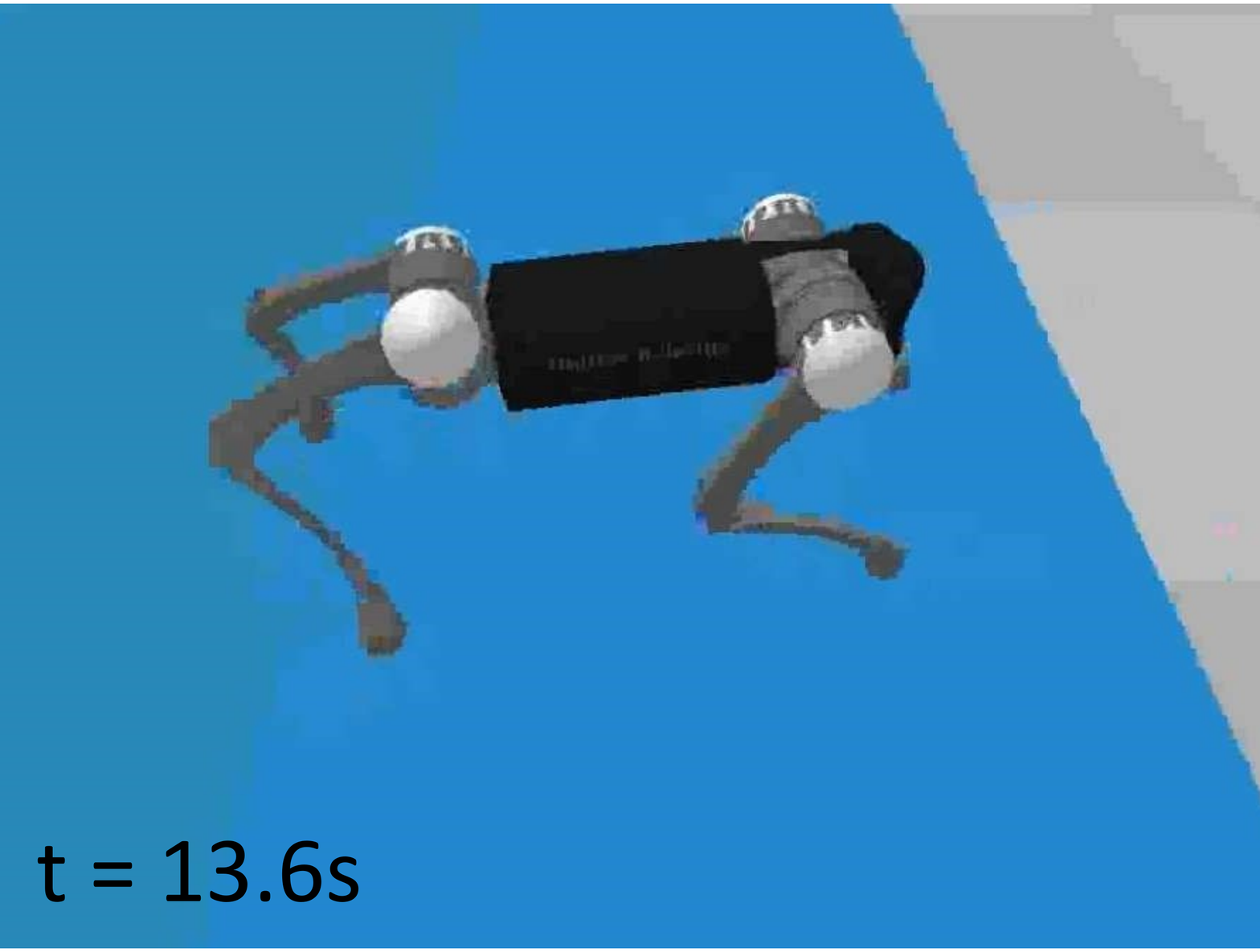}
\includegraphics[width=0.15\linewidth]{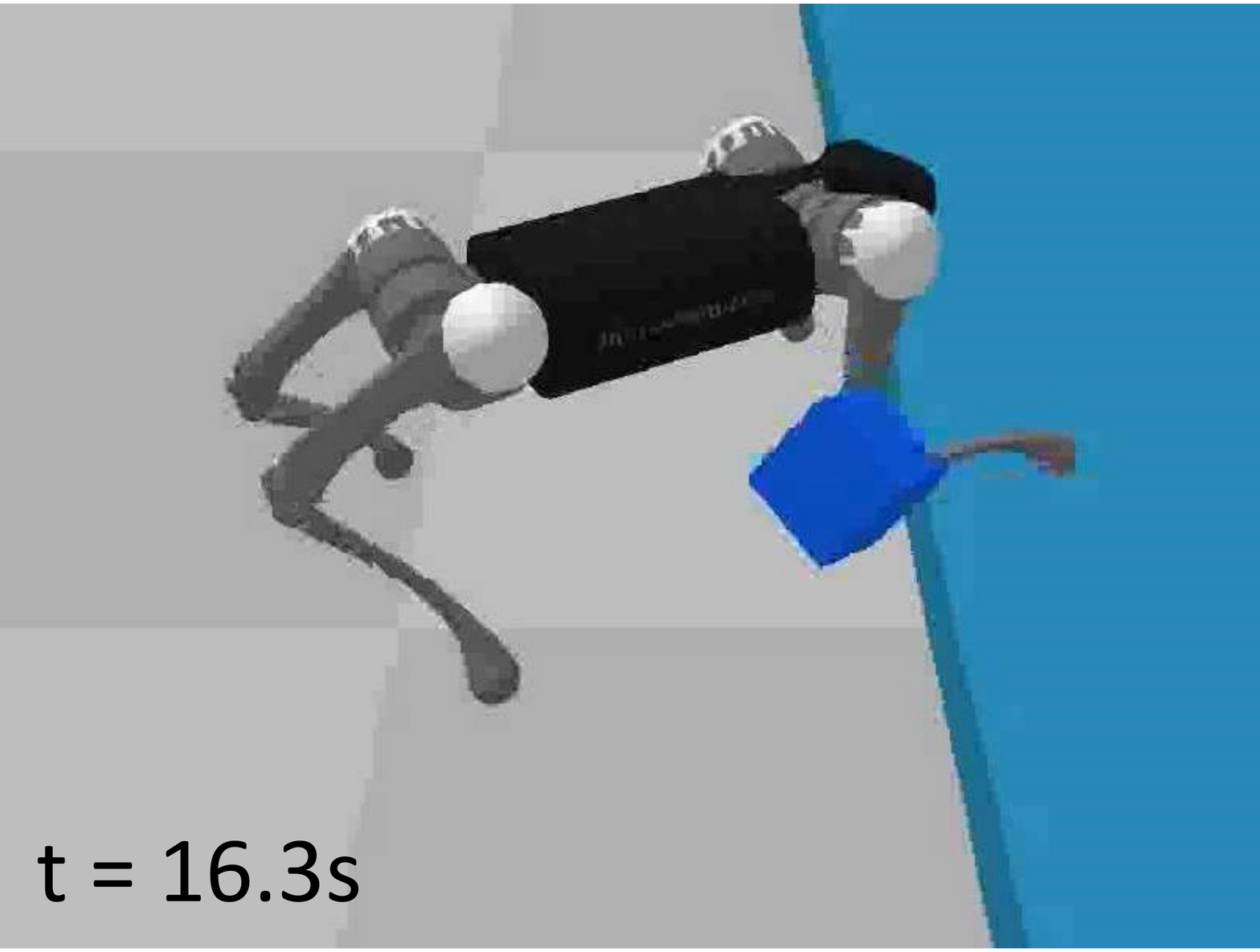}
\includegraphics[width=0.15\linewidth]{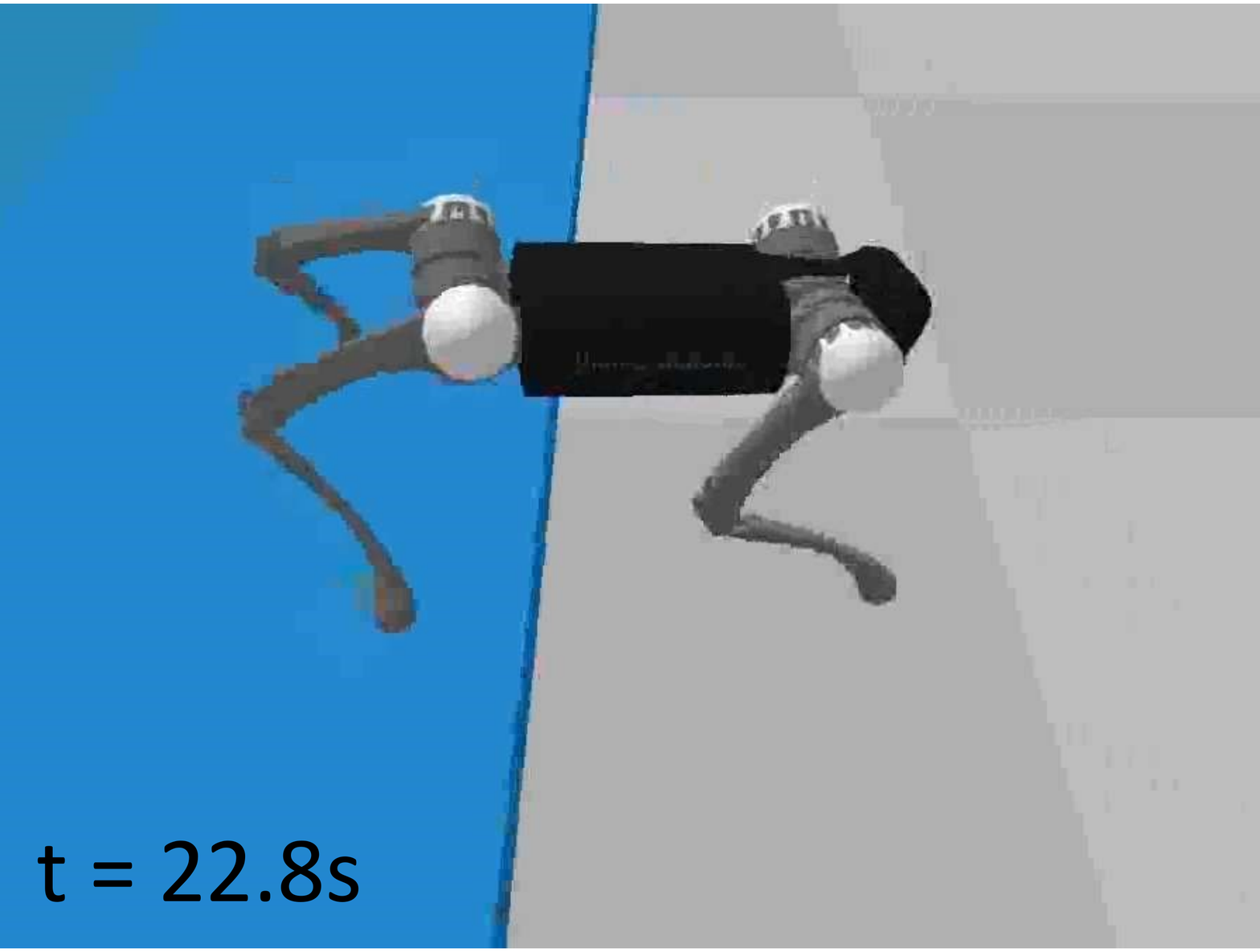}
\includegraphics[width=1.0\linewidth]{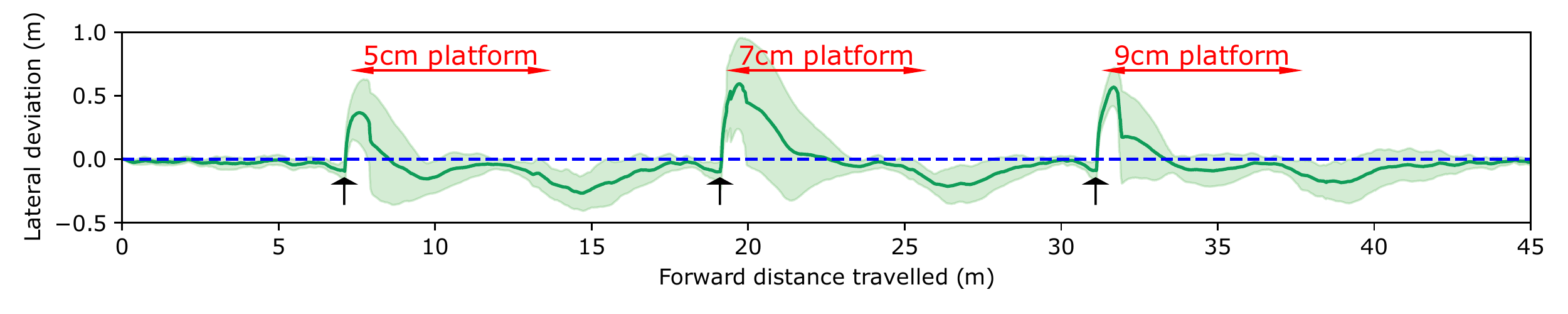}
\includegraphics[width=1.0\linewidth]{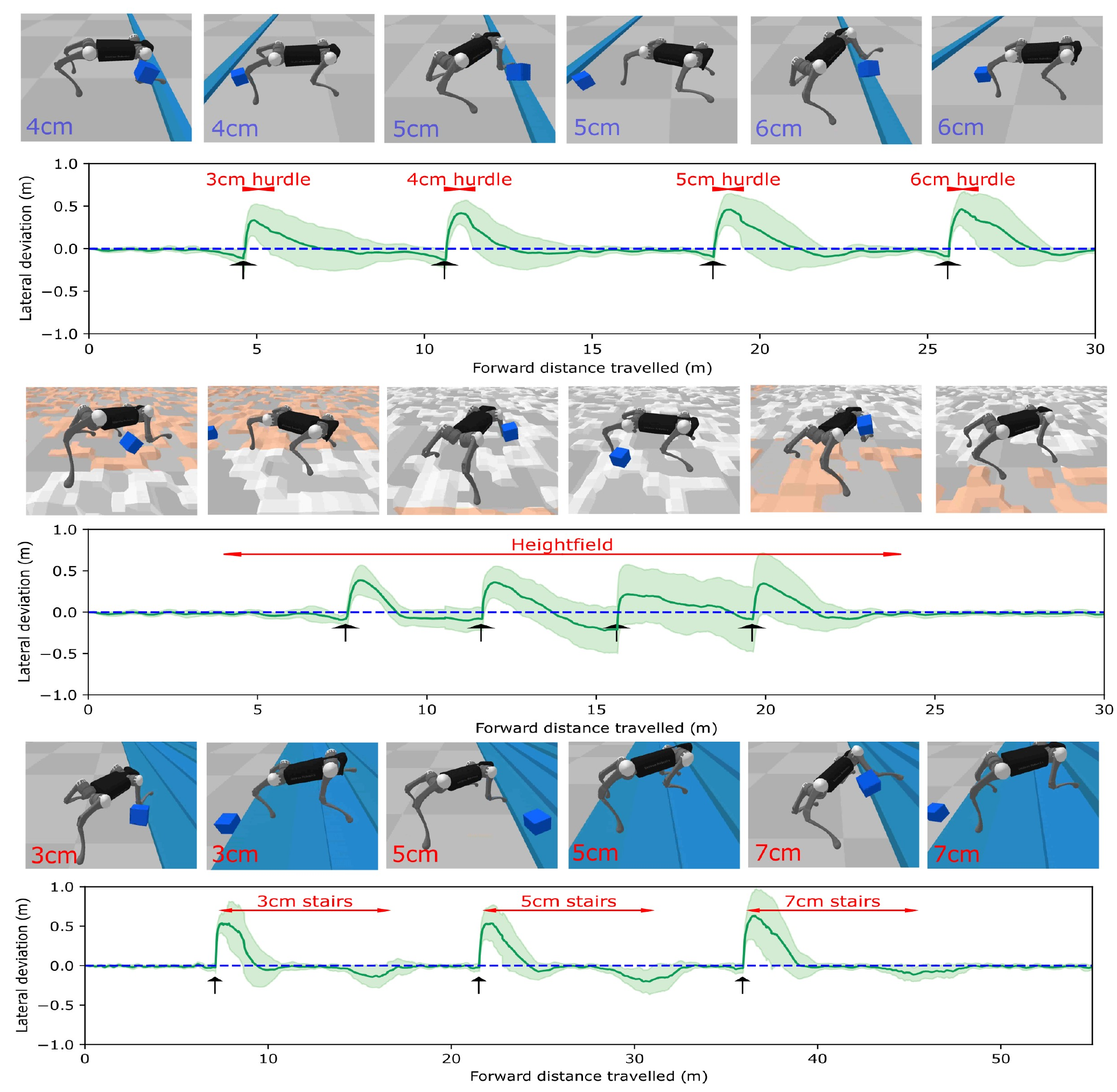}
\vspace{-1mm}
\caption{Robust perceptive locomotion over various terrains by withstanding external force impacts. Black arrows indicate time and direction of the flying objects. In all test cases, the learned policy robustly recovers back to the reference path.}
\vspace{-1mm}
\label{fig:results-robustness-thrown-obj}
\end{figure}

Moreover, we demonstrate robustness of our learned policy against unexpected external collisions. We initialized arbitrary objects of 3.0 kg moving at 10.0 m/s to impact the robot laterally. The impacts occur at the moment before the robot encounters terrain  obstacles, increasing the difficulty of terrain negotiation. Despite these perturbations, our policy robustly recovers from the impact and returns to the reference path after being pushed away. Fig. \ref{fig:results-robustness-thrown-obj} shows the statistical results achieved across 20 unique seeds on platforms. Without perturbations, the learned policy can achieve $100\%$ success rate across all terrains. With perturbation, it achieved $85\%$, $100\%$, $90\%$ completion rate on {\textit{Platforms}}, {\textit{Hurdles}}, {\textit{Heightfield}} respectively.  

Though our work is simulation-only, nonetheless, we argue that sim-to-real deployment is theoretically feasible. Sim-to-real methods for legged locomotion have been extensively reported in the literature, such as domain randomization \cite{peng_learning_2020}, rapid motor adaptation \cite{kumar_rapid_2021}, and fine-tuning in the real world \cite{smith_finetuning_2021}. All of these are compatible with the CPG-guided learning framework that we present in this paper. 

To investigate whether our policy transfers well to realistic scenarios, we tested the policy with additional levels of sensor noise over flat ground. Results are presented in Table \ref{tab:noise}. We note that velocity error and foot contact similarity does not degrade significantly even as noise level increases up to 10x the nominal noise level, demonstrating robustness to random sensor noise. 
The nominal noise level for the heightmap sensor is determined to be 0.005 m. This is obtained by multiplying 2\% nominal depth accuracy with the estimated 0.25 m average robot base height. 

\begin{table}
    \begin{subfigure}[b]{0.40\linewidth}
        \centering
        \begin{tabular}{|l|c|c|}
            \hline
            Noise & $v_{err}$ & $s$ \\ 
            \hline
            1x & 0.10 & 0.88 \\ 
            \hline
            5x & 0.16 & 0.86 \\ 
            \hline
            10x & 0.12 & 0.83 \\ 
            \hline
            25x & 0.37 & 0.69 \\ 
            \hline
        \end{tabular}
    \end{subfigure}
    \hfill
    \begin{subfigure}[b]{0.58\linewidth}
        \centering
        \begin{tabular}{|l|c|}
            \hline
            Parameter & 1x Noise std. dev \\ 
            \hline
            Motor Position & 0.01 rad \\ 
            \hline
            Base RPY & 0.01 rad \\ 
            \hline
            Base ang. vel. & 0.1 rad/s \\ 
            \hline
            Heightmap rays & 0.005 m \\ 
            \hline
        \end{tabular}
    \end{subfigure}
\caption{Evaluation noise levels: \label{tab:noise}(a) Gait metrics at varying levels of noise. $v_{err}$ is RMSE in base velocity, and $s$ denotes Hamming similarity; \label{tab:noise_dist}(b) Noise distributions at the 1x-noise level. Each is a zero-mean Gaussian with the given standard deviation.}
\vspace{-6mm}
\end{table}

\section{Conclusion and Future Work}


This work presents a learning framework to train perceptive locomotion which combines the proprioceptive and exteroceptive data, and task-level enforcement of gait frequency and velocity with natural transitions. This offers high-level inputs to modulate gait paces for terrain-adaptive locomotion. Specifically, the perceptive module uses the latent representation of exteroceptive data effectively, and demonstrates robust tracking of reference gait frequencies and velocities across different terrains. The results show that during the ``entry phase'' of active negotiation with uneven terrains, the front-facing visual perception is more critical, as rear legs can handle steps blindly reasonably well using proprioceptive data. Moreover, the learned policy is robust to unseen external perturbations which were not trained before. 


However, several limitations remain. Firstly, terrain negotiation can be further improved. In the benchmarking test ($2.0$ Hz gait frequency, $0.6$ m/s target velocity), given the same training parameters, our current policy has different negotiation ability across different terrains: traversing steps and hurdles up to 11 cm, stairs terrain up to 9 cm, and heightfield terrains with perturbation up to 8 cm. In future work, additional training time and an improved training curriculum may enable even better terrain negotiation ability.

Secondly, the field of view from visual perception is only for the front legs and not the rear ones. In the current setup, the policy has no information about the terrain around the rear legs. The reason for such a camera configuration is that nature shows that animals are able to negotiate terrains with only front-seeing eyes, so we aim to keep the system with simplicity. Within this work, we kept the exteroceptive heightmap at a straightforward setting with low dimension, such that the rear legs rely on proprioceptive information to blindly step over hurdles. Clearly, this simplicity trades off the completeness of information and can be improved by adding a back camera. In future work, including an additional rear heightmap may improve the policy's robustness and performance.

\balance
\bibliography{main}
\bibliographystyle{IEEEtran}

\end{document}